\def\year{2017}\relax
\documentclass[letterpaper]{article}
\usepackage{aaai17}
\usepackage{times}
\usepackage{helvet}
\usepackage{courier}
\usepackage{times,amsmath,amsfonts,stmaryrd,amsthm,amssymb,enumitem,booktabs}
\usepackage{amsthm}

\newtheorem{lemma}{Lemma}
\usepackage{graphicx}
\DeclareMathOperator*{\argmax}{argmax}
\DeclareMathOperator*{\argmin}{argmin}
\usepackage{color}

\newcommand{\dotprod}[1]{\langle #1 \rangle}
\newcommand{\cD}{\mathcal{D}}
\newcommand{\reals}{\mathbb{R}}

\newcommand{\sign}{\textrm{sign}}
\renewcommand{\eqref}[1]{Eq.~(\ref{#1})}
\newcommand{\figref}[1]{Figure \ref{#1}}
\newcommand{\tabref}[1]{Table \ref{#1}}        
\newcommand{\secref}[1]{Section \ref{#1}}
\newcommand{\appref}[1]{Appendix \ref{#1}}
\newcommand{\p}{\partial}
\newcommand{\ourtitle}{Tunable Sensitivity to Large Errors in Neural Network Training}
\title{\ourtitle}

\author{
Gil Keren \\ Chair of Complex and Intelligent systems \\ University of Passau\\ Passau, Germany \\ gil.keren@uni-passau.de
\And Sivan Sabato \\ Department of Computer Science \\ Ben-Gurion University of the Negev \\ Beer Sheva, Israel 
\And Bj{\"o}rn Schuller \\ Chair of Complex and Intelligent systems \\ University of Passau \\ Passau, Germany, \\ Machine Learning Group \\ Imperial College London, U.K.}

\begin{document}
\maketitle

\newcommand{\tabone}{
\begin{table*}[t]
\caption{Experiment results for single-layer networks}
\begin{center}
\begin{small}
\begin{sc}
\begin{tabular}{|ccc|c|cc|cc|}

\hline

& & &  & \multicolumn{2}{c|}{Test Error} & \multicolumn{2}{c|}{Test Cross-Entropy Loss}\\
Dataset & Layer Size & Momentum & Selected $k$ & $k=1$ & Selected $k$ & $k=1$ & Selected $k$ \\
\hline

 %200 & 0.5  	& 0.5 & 1.86\%				& 1.86\% 			& \textbf{0.088} & 0.234  \\ \hline
MNIST & 400 & 0.5  	& 0.5 & 1.76\% 				& \textbf{1.74\%} 	& \textbf{0.078} & 0.167  \\ \hline
 %600 & 0.5  	& 0.5 & 1.70\% 				& 1.70\% 			& \textbf{0.076} & 0.156  \\ \hline
MNIST & 800 & 0.5  	& 0.5 & 1.67\% 				& \textbf{1.65\%} 	& \textbf{0.072} & 0.150  \\ \hline
MNIST & 1100 & 0.5  	& 0.5 & 1.67\% 				& \textbf{1.65\%} 	& \textbf{0.071} & 0.145  \\ \hline
 %1400 & 0.5  & 0.5 & 1.72\% 				& \textbf{1.63\%} 	&  &   \\ \hline

% 200 & 0.5 & 0.25 		& 17.46\% 				& \textbf{16.83\%} 	& \textbf{0.710} & 1.761  \\ \hline
SVHN & 400 & 0.5 & 0.25 		& 16.88\% 			& \textbf{16.16\%} 	& \textbf{0.661} & 1.576  \\ \hline
% 600 & 0.5 & 0.125 		& 16.36\% 			& \textbf{15.85\%} 	& \textbf{0.656} & 2.954  \\ \hline
SVHN & 800 & 0.5 & 0.125 		& 16.09\% 			& \textbf{15.64\%} 	& \textbf{0.648} & 3.108  \\ \hline
SVHN & 1100 & 0.5 & 0.25 		& 16.04\% 			& \textbf{15.53\%} 	& \textbf{0.626} & 1.525  \\ \hline
% 1400 & 0.5 & 0.125 		& 15.89\% 			& \textbf{15.44\%} 	&  &   \\ \hline

% 200 & 0.5 & 0.125 	& 49.60\% 	& \textbf{47.83\%} 	& \textbf{1.516} & 5.461  \\ \hline
CIFAR-10 & 400 & 0.5 & 0.25 		& 48.32\% 	& \textbf{47.06\%} 	& \textbf{1.430} & 3.034  \\ \hline
% 600 & 0.5 & 0.125 	& 47.68\% 	& \textbf{46.49\%} 	& \textbf{1.437} & 5.501  \\ \hline
CIFAR-10 & 800 & 0.5 & 0.125 	& 46.91\% 	& \textbf{46.01\%} 	& \textbf{1.388} & 5.645  \\ \hline
CIFAR-10 & 1100 & 0.5 & 0.25 	& 46.43\%  	& \textbf{45.84\%} 	& \textbf{1.410} & 2.820  \\ \hline
% 1400 & 0.5 & 0.25 	& 46.29\% 	& \textbf{45.83\%} 	&  &   \\ \hline

CIFAR-100 & 400 & 0.5 & 0.25 		& 75.18\% 	& \textbf{74.41\%} 	& \textbf{3.302} & 6.931  \\ \hline
CIFAR-100 & 800 & 0.5 & 0.25 		& 74.04\% 	& \textbf{73.78\%} 	& \textbf{3.260} & 7.449  \\ \hline
CIFAR-100 & 1100 & 0.5 & 0.125 		& 73.69\% 	& \textbf{73.11\%} 	& \textbf{3.239} & 13.557  \\ \hline
\end{tabular}
\end{sc}
\end{small}

\label{tabOneLayer}
\end{center}
\end{table*}
}

\newcommand{\taboneSup}{
\begin{table*}[t]
\caption{Experiment results for single-layer networks}
\begin{center}
\begin{small}
\begin{sc}
\begin{tabular}{|ccc|c|cc|cc|}
\hline

& & &  & \multicolumn{2}{c|}{Test Error} & \multicolumn{2}{c|}{Test Cross-Entropy Loss}\\
Dataset & Layer Size & Momentum & Selected $k$ & $k=1$ & Selected $k$ & $k=1$ & Selected $k$ \\
\hline
MNIST & 400 & 0 		& 0.5 & 1.71\% 				& \textbf{1.70\%} 	& \textbf{0.0757} & 0.148  \\ \hline
MNIST & 800 & 0  	& 0.5 & \textbf{1.66\%} 		& 1.67\% 			& \textbf{0.070} & 0.137  \\ \hline
MNIST & 1100 & 0 	& 0.5 &  1.64\% 				& \textbf{1.62\%} 	& \textbf{0.068} & 0.131  \\ \hline
MNIST & 400 & 0.9  	& 0.5 & 1.75\% 				& 1.75\% 			& \textbf{0.073} & 0.140  \\ \hline
MNIST & 800 & 0.9  	& 2 & 1.71\% 				& \textbf{1.63\%} 	& 0.070 &  \textbf{0.054} \\ \hline
MNIST & 1100 & 0.9  & 0.5 & 1.74\% 				& \textbf{1.69\%} 	& \textbf{0.069} & 0.127  \\ \hline
 
SVHN & 400 & 0 & 0.25 			& 16.84\% 			& \textbf{16.09\%} 	& \textbf{0.658} & 1.575  \\ \hline
SVHN & 800 & 0 & 0.25 			& 16.19\% 			& \textbf{15.71\%} 	& \textbf{0.641} & 1.534  \\ \hline
SVHN & 1100 & 0 & 0.25 			& 15.97\%			& \textbf{15.68\%} 	& \textbf{0.636} & 1.493  \\ \hline
SVHN & 400 & 0.9 & 0.125 		& 16.65\% 			& \textbf{16.30\%} 	& \textbf{0.679} & 2.861  \\ \hline
SVHN & 800 & 0.9 & 0.25 		& 16.15\% 			& \textbf{15.68\%} 	& \textbf{0.675} & 1.632  \\ \hline
SVHN & 1100 & 0.9 & 0.25 		& 15.85\% 			& \textbf{15.47\%} 	& \textbf{0.640} & 1.657  \\ \hline

CIFAR-10 & 400 & 0 & 0.125 		& 48.15\% 	& \textbf{46.91\%} 	& \textbf{1.435} & 5.609  \\ \hline
CIFAR-10 & 800 & 0 & 0.125 		& 46.92\% 	& \textbf{46.14\%} 	& \textbf{1.390} & 5.390  \\ \hline
CIFAR-10 & 1100 & 0 & 0.125 		& 46.63\% 	& \textbf{46.00\%} 	& \textbf{1.356} & 5.290  \\ \hline
CIFAR-10 & 400 & 0.9 & 0.0625 	& 48.19\% 	& \textbf{46.71\%} 	& \textbf{1.518} & 11.049  \\ \hline
CIFAR-10 & 800 & 0.9 & 0.125 	& 47.09\% 	& \textbf{46.16\%} 	& \textbf{1.616} & 5.294  \\ \hline
CIFAR-10 & 1100 & 0.9 & 0.125 	& 46.71\% 	& \textbf{45.77\%} 	& \textbf{1.850} & 5.904  \\ \hline

CIFAR-100 & 400 & 0.9 & 0.25 		& 74.96\% 	& \textbf{74.28\%} 	& \textbf{3.306} & 7.348  \\ \hline
CIFAR-100 & 800 & 0.9 & 0.125 		& 74.12\% 	& \textbf{73.47\%} 	& \textbf{3.327} & 13.267  \\ \hline
CIFAR-100 & 1100 & 0.9 & 0.25 		& 73.47\% 	& \textbf{73.19\%} 	& \textbf{3.235} & 7.489  \\ \hline

\end{tabular}
\end{sc}
\end{small}

\label{tabOneLayerSup}
\end{center}
\end{table*}
}

\newcommand{\tabonemnist}{
\begin{table*}[t]
\caption{Experiment results for single-layer networks, the MNIST dataset}
\begin{center}
\begin{small}
\begin{sc}
\begin{tabular}{|cc|c|cc|cc|}
\hline

& & & \multicolumn{2}{c|}{Test Error} & \multicolumn{2}{c|}{Test Cross-Entropy Loss}\\
Layer Size & Momentum & Selected $k$ & $k=1$ & Selected $k$ & $k=1$ & Selected $k$ \\
\hline

 %200 & 0.5  	& 0.5 & 1.86\%				& 1.86\% 			& \textbf{0.088} & 0.234  \\ \hline
 400 & 0.5  	& 0.5 & 1.76\% 				& \textbf{1.74\%} 	& \textbf{0.078} & 0.167  \\ \hline
 %600 & 0.5  	& 0.5 & 1.70\% 				& 1.70\% 			& \textbf{0.076} & 0.156  \\ \hline
 800 & 0.5  	& 0.5 & 1.67\% 				& \textbf{1.65\%} 	& \textbf{0.072} & 0.150  \\ \hline
 1100 & 0.5  	& 0.5 & 1.67\% 				& \textbf{1.65\%} 	& \textbf{0.071} & 0.145  \\ \hline
 %1400 & 0.5  & 0.5 & 1.72\% 				& \textbf{1.63\%} 	&  &   \\ \hline

\end{tabular}
\end{sc}
\end{small}

\label{tabOneLayerMNIST}
\end{center}
\end{table*}
}

\newcommand{\tabonemnistSup}{
\begin{table*}[t]
\caption{Experiment results for single-layer networks, the MNIST dataset}
\begin{center}
\begin{small}
\begin{sc}
\begin{tabular}{|cc|c|cc|cc|}
\hline

& & & \multicolumn{2}{c|}{Test Error} & \multicolumn{2}{c|}{Test Cross-Entropy Loss}\\
Layer Size & Momentum & Selected $k$ & $k=1$ & Selected $k$ & $k=1$ & Selected $k$ \\
\hline
 %200 & 0 		& 0.5 & 1.87\% 				& \textbf{1.85\%} 	& \textbf{0.084} & 0.189  \\ \hline
 400 & 0 		& 0.5 & 1.71\% 				& \textbf{1.70\%} 	& \textbf{0.0757} & 0.148  \\ \hline
 %600 & 0  	& 0.5 & \textbf{1.64\%} 		& 1.65\% 			& \textbf{0.072} & 0.138  \\ \hline
 800 & 0  	& 0.5 & \textbf{1.66\%} 		& 1.67\% 			& \textbf{0.070} & 0.137  \\ \hline
 1100 & 0 	& 0.5 &  1.64\% 				& \textbf{1.62\%} 	& \textbf{0.068} & 0.131  \\ \hline
 %1400 & 0  	& 0.5 &  1.66\% 				& \textbf{1.61\%} 	&  &   \\ \hline

 %200 & 0.9  	& 0.5 & 1.91\% 				& \textbf{1.84\%} 	& \textbf{0.082} & 0.165  \\ \hline
 400 & 0.9  	& 0.5 & 1.75\% 				& 1.75\% 			& \textbf{0.073} & 0.140  \\ \hline
 %600 & 0.9  	& 0.5 & 1.75\% 				& \textbf{1.70\%} 	& \textbf{0.071} & 0.137  \\ \hline
 800 & 0.9  	& 2 & 1.71\% 				& \textbf{1.63\%} 	& 0.070 &  \textbf{0.054} \\ \hline
 1100 & 0.9  & 0.5 & 1.74\% 				& \textbf{1.69\%} 	& \textbf{0.069} & 0.127  \\ \hline
 %1400 & 0.9  & 0.5 & 1.72\% 				& \textbf{1.68\%} 	&  &   \\ \hline
\end{tabular}
\end{sc}
\end{small}

\label{tabOneLayerMNISTSup}
\end{center}
\end{table*}
}

\newcommand{\tabonesvhn}{
\begin{table*}[t]
\caption{Experiment results for single-layer networks, the SVHN dataset}
\begin{center}
\begin{small}
\begin{sc}
\begin{tabular}{|cc|c|cc|cc|}
\hline

 & & & \multicolumn{2}{c|}{Test Error} & \multicolumn{2}{c|}{Test Cross-Entropy Loss}\\
Layer Size & Momentum & Selected $k$ & $k=1$ & Selected $k$ & $k=1$ & Selected $k$ \\
\hline

% 200 & 0.5 & 0.25 		& 17.46\% 				& \textbf{16.83\%} 	& \textbf{0.710} & 1.761  \\ \hline
 400 & 0.5 & 0.25 		& 16.88\% 			& \textbf{16.16\%} 	& \textbf{0.661} & 1.576  \\ \hline
% 600 & 0.5 & 0.125 		& 16.36\% 			& \textbf{15.85\%} 	& \textbf{0.656} & 2.954  \\ \hline
 800 & 0.5 & 0.125 		& 16.09\% 			& \textbf{15.64\%} 	& \textbf{0.648} & 3.108  \\ \hline
 1100 & 0.5 & 0.25 		& 16.04\% 			& \textbf{15.53\%} 	& \textbf{0.626} & 1.525  \\ \hline
% 1400 & 0.5 & 0.125 		& 15.89\% 			& \textbf{15.44\%} 	&  &   \\ \hline

\end{tabular}
\end{sc}
\end{small}

\label{tabOneLayerSVHN}
\end{center}
\end{table*}
}

\newcommand{\tabonesvhnSup}{
\begin{table*}[t]
\caption{Experiment results for single-layer networks, the SVHN dataset}
\begin{center}
\begin{small}
\begin{sc}
\begin{tabular}{|cc|c|cc|cc|}
\hline

 & & & \multicolumn{2}{c|}{Test Error} & \multicolumn{2}{c|}{Test Cross-Entropy Loss}\\
Layer Size & Momentum & Selected $k$ & $k=1$ & Selected $k$ & $k=1$ & Selected $k$ \\
\hline
% 200 & 0 & 0.125 			& 17.42\% 			& \textbf{16.60\%} 	& \textbf{0.702} & 3.354  \\ \hline
 400 & 0 & 0.25 			& 16.84\% 			& \textbf{16.09\%} 	& \textbf{0.658} & 1.575  \\ \hline
% 600 & 0 & 0.125 			& 16.41\% 			& \textbf{15.85\%} 	& \textbf{0.675} & 2.925  \\ \hline
 800 & 0 & 0.25 			& 16.19\% 			& \textbf{15.71\%} 	& \textbf{0.641} & 1.534  \\ \hline
 1100 & 0 & 0.25 			& 15.97\%			& \textbf{15.68\%} 	& \textbf{0.636} & 1.493  \\ \hline
% 1400 & 0 & 0.125 		& 15.98\% 				& \textbf{15.73\%} 	&  &   \\ \hline

% 200 & 0.9 & 0.125 		& 17.39\% 			& \textbf{16.58\%} 	& \textbf{0.710} & 3.437  \\ \hline
 400 & 0.9 & 0.125 		& 16.65\% 			& \textbf{16.30\%} 	& \textbf{0.679} & 2.861  \\ \hline
% 600 & 0.9 & 0.125 		& 16.33\% 			& \textbf{15.67\%} 	& \textbf{0.661} & 3.084  \\ \hline
 800 & 0.9 & 0.25 		& 16.15\% 			& \textbf{15.68\%} 	& \textbf{0.675} & 1.632  \\ \hline
 1100 & 0.9 & 0.25 		& 15.85\% 			& \textbf{15.47\%} 	& \textbf{0.640} & 1.657  \\ \hline
% 1400 & 0.9 & 0.25 		& 15.77\% 			& \textbf{15.44\%} 	&  &   \\ \hline
\end{tabular}
\end{sc}
\end{small}

\label{tabOneLayerSVHNSup}
\end{center}
\end{table*}
}

\newcommand{\tabonecifarten}{
\begin{table*}[t]
\caption{Experiment results for single-layer networks, the CIFAR-10 dataset}
\begin{center}
\begin{small}
\begin{sc}
\begin{tabular}{|cc|c|cc|cc|}
\hline

& & & \multicolumn{2}{c|}{Test Error} & \multicolumn{2}{c|}{Test Cross-Entropy Loss}\\
Layer Size & Momentum & Selected $k$ & $k=1$ & Selected $k$ & $k=1$ & Selected $k$ \\
\hline

% 200 & 0.5 & 0.125 	& 49.60\% 	& \textbf{47.83\%} 	& \textbf{1.516} & 5.461  \\ \hline
 400 & 0.5 & 0.25 		& 48.32\% 	& \textbf{47.06\%} 	& \textbf{1.430} & 3.034  \\ \hline
% 600 & 0.5 & 0.125 	& 47.68\% 	& \textbf{46.49\%} 	& \textbf{1.437} & 5.501  \\ \hline
 800 & 0.5 & 0.125 	& 46.91\% 	& \textbf{46.01\%} 	& \textbf{1.388} & 5.645  \\ \hline
 1100 & 0.5 & 0.25 	& 46.43\%  	& \textbf{45.84\%} 	& \textbf{1.410} & 2.820  \\ \hline
% 1400 & 0.5 & 0.25 	& 46.29\% 	& \textbf{45.83\%} 	&  &   \\ \hline

%\belowspace
\end{tabular}
\end{sc}
\end{small}

\label{tabOneLayerCIFAR10}
\end{center}
\end{table*}
}

\newcommand{\tabonecifartenSup}{
\begin{table*}[t]
\caption{Experiment results for single-layer networks, the CIFAR-10 dataset}
\begin{center}
\begin{small}
\begin{sc}
\begin{tabular}{|cc|c|cc|cc|}
\hline

& & & \multicolumn{2}{c|}{Test Error} & \multicolumn{2}{c|}{Test Cross-Entropy Loss}\\
Layer Size & Momentum & Selected $k$ & $k=1$ & Selected $k$ & $k=1$ & Selected $k$ \\
\hline
% 200 & 0 & 0.0625 		& 49.79\% 	& \textbf{47.78\%} 	& \textbf{1.449} & 10.798  \\ \hline
 400 & 0 & 0.125 		& 48.15\% 	& \textbf{46.91\%} 	& \textbf{1.435} & 5.609  \\ \hline
% 600 & 0 & 0.125 		& 47.59\% 	& \textbf{46.28\%} 	& \textbf{1.383} & 5.594  \\ \hline
 800 & 0 & 0.125 		& 46.92\% 	& \textbf{46.14\%} 	& \textbf{1.390} & 5.390  \\ \hline
 1100 & 0 & 0.125 		& 46.63\% 	& \textbf{46.00\%} 	& \textbf{1.356} & 5.290  \\ \hline
% 1400 & 0 & 0.125 		& 46.14\% 	& \textbf{45.81\%} 	&  &   \\ \hline

% 200 & 0.9 & 0.125 	& 49.56\% 	& \textbf{47.83\%} 	& \textbf{1.510} & 5.809  \\ \hline
 400 & 0.9 & 0.0625 	& 48.19\% 	& \textbf{46.71\%} 	& \textbf{1.518} & 11.049  \\ \hline
% 600 & 0.9 & 0.125 	& 47.73\% 	& \textbf{46.19\%} 	& \textbf{1.521} & 5.810  \\ \hline
 800 & 0.9 & 0.125 	& 47.09\% 	& \textbf{46.16\%} 	& \textbf{1.616} & 5.294  \\ \hline
 1100 & 0.9 & 0.125 	& 46.71\% 	& \textbf{45.77\%} 	& \textbf{1.850} & 5.904  \\ \hline
% 1400 & 0.9 & 0.125 	& 46.29\% 	& \textbf{45.81\%} 	&  &   \\ \hline
%\belowspace
\end{tabular}
\end{sc}
\end{small}

\label{tabOneLayerCIFAR10Sup}
\end{center}
\end{table*}
}

\newcommand{\tabonecifarhundred}{
\begin{table*}[t]
\caption{Experiment results for single-layer networks, the CIFAR-100 dataset}
\begin{center}
\begin{small}
\begin{sc}
\begin{tabular}{|cc|c|cc|cc|}
\hline

& & & \multicolumn{2}{c|}{Test Error} & \multicolumn{2}{c|}{Test Cross-Entropy Loss}\\
Layer Size & Momentum & Selected $k$ & $k=1$ & Selected $k$ & $k=1$ & Selected $k$ \\
\hline
 400 & 0.5 & 0.25 		& 75.18\% 	& \textbf{74.41\%} 	& \textbf{3.302} & 6.931  \\ \hline
 800 & 0.5 & 0.25 		& 74.04\% 	& \textbf{73.78\%} 	& \textbf{3.260} & 7.449  \\ \hline
 1100 & 0.5 & 0.125 		& 73.69\% 	& \textbf{73.11\%} 	& \textbf{3.239} & 13.557  \\ \hline

%\belowspace
\end{tabular}
\end{sc}
\end{small}

\label{tabOneLayerCIFAR100}
\end{center}
\end{table*}
}

\newcommand{\tabonecifarhundredSup}{
\begin{table*}[t]
\caption{Experiment results for single-layer networks, the CIFAR-100 dataset}
\begin{center}
\begin{small}
\begin{sc}
\begin{tabular}{|cc|c|cc|cc|}
\hline

& & & \multicolumn{2}{c|}{Test Error} & \multicolumn{2}{c|}{Test Cross-Entropy Loss}\\
Layer Size & Momentum & Selected $k$ & $k=1$ & Selected $k$ & $k=1$ & Selected $k$ \\
\hline
 400 & 0.9 & 0.25 		& 74.96\% 	& \textbf{74.28\%} 	& \textbf{3.306} & 7.348  \\ \hline
 800 & 0.9 & 0.125 		& 74.12\% 	& \textbf{73.47\%} 	& \textbf{3.327} & 13.267  \\ \hline
 1100 & 0.9 & 0.25 		& 73.47\% 	& \textbf{73.19\%} 	& \textbf{3.235} & 7.489  \\ \hline

%\belowspace
\end{tabular}
\end{sc}
\end{small}

\label{tabOneLayerCIFAR100Sup}
\end{center}
\end{table*}
}

\newcommand{\tabthree}{
\begin{table*}[t]
\caption{Experiment results for 3-layer networks}
\begin{center}
\begin{small}
\begin{sc}
\begin{tabular}{|ccc|c|cc|cc|}
\hline

& & & & \multicolumn{2}{c|}{Test Error} & \multicolumn{2}{c|}{Test CE Loss}\\
Dataset & Layer Sizes & Mom' & Selected $k$ & $k=1$ & Selected $k$ & $k=1$ & Selected $k$ \\
\hline
MNIST 	& 400 & 0.5 		& 1	 	& ---			& --- 				& --- 			 & ---  \\ \hline
MNIST 	& 800 & 0.5 		& 1		& ---			& --- 				& --- 			 & ---  \\ \hline
SVHN  	& 400 & 0.5 		& 2	 	& 16.52\% 		& 16.52\% 			& 1.604			 & \textbf{0.968}  \\ \hline
SVHN  	& 800 & 0.5 		& 1		& ---			& --- 				& --- 			 & ---  \\ \hline
CIFAR-10 & 400 & 0.5 	& 2	 	& 46.81\% 		& \textbf{46.63\%} 	& 3.023			 & \textbf{2.121}  \\ \hline
CIFAR-10 & 800 & 0.5 	& 1		& ---			& --- 				& --- 			 & ---  \\ \hline
CIFAR-100 & 400 & 0.5 	& 0.5 	& 75.20\% 		& \textbf{74.95\%} 	& \textbf{3.378}	 & 4.511  \\ \hline
CIFAR-100 & 800 & 0.5 	& 1		& ---			& --- 				& --- 			 & ---  \\ \hline
\end{tabular}
\end{sc}
\end{small}

\label{tab3layer}
\end{center}
\end{table*}
}

\newcommand{\tabthreeSup}{
\begin{table*}[t]
\caption{Experiment results for 3-layer networks}
\begin{center}
\begin{small}
\begin{sc}
\begin{tabular}{|ccc|c|cc|cc|}
\hline

& & & & \multicolumn{2}{c|}{Test Error} & \multicolumn{2}{c|}{Test CE Loss}\\
Dataset & Layer Sizes & Mom' & Selected $k$ & $k=1$ & Selected $k$ & $k=1$ & Selected $k$ \\
\hline
MNIST 	& 400 & 0 		& 1	 	& ---			& --- 				& --- 			 & ---  \\ \hline
MNIST 	& 800 & 0 		& 1		& ---			& --- 				& --- 			 & ---  \\ \hline
MNIST 	& 400 & 0.9 		& 1	 	& ---			& --- 				& --- 			 & ---  \\ \hline
MNIST 	& 800 & 0.9 		& 0.5	& 1.60\% 		& \textbf{1.53\%}	& \textbf{0.091}	 & 0.189  \\ \hline
SVHN  	& 400 & 0.9 		& 1	 	& ---			& --- 				& --- 			 & ---  \\ \hline
SVHN  	& 800 & 0.9 		& 2		& 16.14\% 		& \textbf{15.96\%} 	& 1.651			 & \textbf{1.062}  \\ \hline
CIFAR-10 & 400 & 0.9 	& 2	 	& 47.52\% 		& \textbf{46.92\%} 	& 2.226			 & \textbf{2.010}  \\ \hline
CIFAR-10 & 800 & 0.9 	& 2		& 45.27\% 		& \textbf{44.26\%}	& 2.855			 & \textbf{2.341}  \\ \hline
CIFAR-100 & 400 & 0.9 	& 0.25 	& 74.97\% 		& \textbf{74.52\%} 	& \textbf{3.356}	 & 8.520  \\ \hline
CIFAR-100 & 800 & 0.9 	& 0.5 	& 74.48\% 		& \textbf{73.17\%} 	& \textbf{4.133}	 & 8.642  \\ \hline
\end{tabular}
\end{sc}
\end{small}

\label{tab3layerSup}
\end{center}
\end{table*}
}

\newcommand{\tabfive}{
\begin{table*}[t]
\caption{Experiment results for 5-layer networks}
\begin{center}
\begin{small}
\begin{sc}
\begin{tabular}{|ccc|c|cc|cc|}
\hline

& & & & \multicolumn{2}{c|}{Test Error} & \multicolumn{2}{c|}{Test CE Loss}\\
Dataset & Layer Sizes & Mom' & Selected $k$ & $k=1$ & Selected $k$ & $k=1$ & Selected $k$ \\
\hline
MNIST 		& 400 	& 0.5 		& 0.5 	& 1.71\%			& \textbf{1.69\%	}		& \textbf{0.113} 	& 0.224 	\\ \hline
MNIST 		& 800	& 0.5 		& 0.25	& 1.61\%			& \textbf{1.60\%	}		& \textbf{0.118} & 0.390  	\\ \hline
SVHN 		& 400 	& 0.5 		& 4	 	& 17.41\%		& \textbf{16.49\%}		& 1.436 	& \textbf{0.708} 	\\ \hline
SVHN 		& 800	& 0.5 		& 0.5	& 17.07\%		& \textbf{16.61\%} 		& \textbf{1.343} 	& 2.604  \\ \hline
CIFAR-10 	& 400 	& 0.5 		& 2	 	& 48.05\%		& \textbf{47.85\%}		& 2.017 	& \textbf{1.962} 	\\ \hline
CIFAR-10 	& 800	& 0.5 		& 4		& \textbf{44.21\%}		& 44.24\% 		& 4.610 	& \textbf{1.677} 	\\ \hline
CIFAR-100 	& 400 	& 0.5 		& 2	 	& 75.69\%		& \textbf{75.48\%}		& 3.611 	& \textbf{3.228} 	\\ \hline
CIFAR-100 	& 800 	& 0.5 		& 2	 	& 74.10\%		& \textbf{73.57\%}		& 4.650 	& \textbf{4.439} 	\\ \hline
MNIST 		& 400 	& 0.9 		& 1	 	& ---			& ---					& ---			 	& --- 	\\ \hline
MNIST 		& 800	& 0.9 		& 4		& \textbf{1.58\%}	& 1.60				& 0.098 & \textbf{0.060}  	\\ \hline
SVHN 		& 400 	& 0.9 		& 4	 	& 17.89\%		& \textbf{16.54\%}		& 1.284 	& \textbf{0.718} 	\\ \hline
SVHN 		& 800 	& 0.9 		& 2	 	& 16.24\%		& \textbf{15.73\%}		& 1.647 	& \textbf{0.998} 	\\ \hline
CIFAR-10 	& 400 	& 0.9 		& 4	 	& 47.91\%		& \textbf{47.57\%} 		& 2.202 	& \textbf{1.648} 	\\ \hline
CIFAR-10 	& 800	& 0.9 		& 2		& 45.69\%		& \textbf{44.11\%} 		& 3.316 	& \textbf{2.171} 	\\ \hline
CIFAR-100 	& 400 	& 0.9 		& 1	 	& ---			& ---					& ---			 	& --- 	\\ \hline
CIFAR-100 	& 800 	& 0.9 		& 4	 	& \textbf{74.32\%}	& 74.62\%			& 3.872 	& \textbf{3.432} 	\\ \hline
%\belowspace

\end{tabular}
\end{sc}
\end{small}

\label{tab5layer}
\end{center}
\end{table*}
}

\newcommand{\tabfiveSup}{
\begin{table*}[t]
\caption{Experiment results for 5-layer networks}
\begin{center}
\begin{small}
\begin{sc}
\begin{tabular}{|ccc|c|cc|cc|}
\hline

& & & & \multicolumn{2}{c|}{Test Error} & \multicolumn{2}{c|}{Test CE Loss}\\
Dataset & Layer Sizes & Mom' & Selected $k$ & $k=1$ & Selected $k$ & $k=1$ & Selected $k$ \\
\hline
MNIST 		& 400 	& 0.9 		& 1	 	& ---			& ---					& ---			 	& --- 	\\ \hline
MNIST 		& 800	& 0.9 		& 4		& \textbf{1.58\%}	& 1.60				& 0.098 & \textbf{0.060}  	\\ \hline
SVHN 		& 400 	& 0.9 		& 4	 	& 17.89\%		& \textbf{16.54\%}		& 1.284 	& \textbf{0.718} 	\\ \hline
SVHN 		& 800 	& 0.9 		& 2	 	& 16.24\%		& \textbf{15.73\%}		& 1.647 	& \textbf{0.998} 	\\ \hline
CIFAR-10 	& 400 	& 0.9 		& 4	 	& 47.91\%		& \textbf{47.57\%} 		& 2.202 	& \textbf{1.648} 	\\ \hline
CIFAR-10 	& 800	& 0.9 		& 2		& 45.69\%		& \textbf{44.11\%} 		& 3.316 	& \textbf{2.171} 	\\ \hline
CIFAR-100 	& 400 	& 0.9 		& 1	 	& ---			& ---					& ---			 	& --- 	\\ \hline
CIFAR-100 	& 800 	& 0.9 		& 4	 	& \textbf{74.32\%}	& 74.62\%			& 3.872 	& \textbf{3.432} 	\\ \hline
%\belowspace

\end{tabular}
\end{sc}
\end{small}

\label{tab5layerSup}
\end{center}
\end{table*}
}

%%% Local Variables:
%%% mode: latex
%%% TeX-master: "cost_change_sivan"
%%% End:

\begin{abstract} 
When humans learn a new concept, they might ignore examples that they cannot make sense of at first, and only later focus on such examples, when they are more useful for learning. We propose incorporating this idea of tunable sensitivity for hard examples in neural network learning, using a new generalization of the cross-entropy gradient step, which can be used in place of the gradient in any gradient-based training method. The generalized gradient is parameterized by a value that controls the sensitivity of the training process to harder training examples. We tested our method on several benchmark datasets. We propose, and corroborate in our experiments, that the optimal level of sensitivity to hard example is positively correlated with the depth of the network. Moreover, the test prediction error obtained by our method is generally lower than that of the vanilla cross-entropy gradient learner. We therefore conclude that tunable sensitivity can be helpful for neural network learning. 

\end{abstract}

%%% Local Variables:
%%% mode: latex
%%% TeX-master: "cost_change"
%%% End:

\section{Introduction} \label{secIntroduction}
In recent years, neural networks have become empirically successful in a wide range of supervised learning applications, such as computer vision \cite{krizhevsky2012imagenet,Szegedy_2015_CVPR}, speech recognition \cite{hinton2012deep}, natural language processing \cite{sutskever2014sequence} and computational paralinguistics \cite{Keren16-CRAa,keren2016convolutional}.
Standard implementations of training feed-forward neural networks for classification are based on gradient-based stochastic optimization, usually optimizing the empirical cross-entropy loss \cite{hinton1989connectionist}.
%\cite{robbins1951stochastic}

However, the cross-entropy is only a surrogate for the true objective of supervised network training, which is in most cases to reduce the probability of a prediction error (or in some case BLEU score, word-error-rate, etc). When optimizing using the cross-entropy loss, as we show below, the effect of training examples on the gradient is linear in the \emph{prediction bias}, which is the difference between the network-predicted class probabilities and the target class probabilities. In particular, a wrong confident prediction induces a larger gradient than a similarly wrong, but less confident, prediction. 

In contrast, humans sometimes employ a different approach to learning: when learning new concepts, they might ignore the examples they feel they do not understand, and focus more on the examples that are more useful to them. When improving proficiency regarding a familiar concept, they might focus on the harder examples, as these can contain more relevant information for the advanced learner. We make a first step towards incorporating this ability into neural network models, by proposing a learning algorithm with a tunable sensitivity to easy and hard training examples. Intuitions about human cognition have often inspired successful machine learning approaches \cite{bengio2009curriculum,cho2015describing,lake2016building}. In this work we show that this can be the case also for tunable sensitivity. 

Intuitively, the depth of the model should be positively correlated with the optimal sensitivity to hard examples. When the network is relatively shallow, its modeling
capacity is limited. In this case, it might be better to reduce sensitivity to
hard examples, since it is likely that these examples cannot be modeled
correctly by the network, and so adjusting the model according to these
examples might only degrade overall prediction accuracy. On the other hand,
when the network is relatively deep, it has a high modeling capacity. In this
case, it might be beneficial to allow more sensitivity to hard examples,
thereby possibly improving the accuracy of the final learned model.

Our learning algorithm works by generalizing the cross-entropy gradient, where the new function can be used instead of the gradient in any gradient-based optimization method for neural networks. Many such training methods have been proposed, including, to name a few, Momentum \cite{polyak1964some}, RMSProp \cite{tieleman2012lecture}, and Adam \cite{kingma2015adamICLR}. The proposed generalization is parameterized by a value $k > 0$, that controls the sensitivity of the training process to hard examples, replacing the fixed dependence of the cross-entropy gradient.
When $k=1$ the proposed update rule is exactly the cross-entropy gradient. 
Smaller values of $k$ decrease the sensitivity during training to hard examples, and larger values of $k$ increase it. 

We report experiments on several benchmark datasets. These experiments show, matching our expectations, that in almost all cases prediction error is improved using large values of $k$ for deep networks, small values of $k$ for shallow networks, and values close to the default $k=1$ for networks of medium depth. They further show that using a tunable sensitivity parameter generally improves the results of learning.

%that when $k$ is selected using cross validation, 
%the test prediction error is improved, even though the cross-entropy loss on the test set is sometimes degraded. Moreover, the value of $k$ chosen by cross-validation is almost always smaller than $1$ for a network with a single hidden layer, and almost always larger or equal to $1$ for deeper networks. This corroborates the intuition above, that optimizing the sensitivity to hard example improves the prediction success of the neural network.

The paper is structured as follows: In \secref{secRelated} related work is discussed. \secref{secSetting} presents our setting and notation. A framework for generalizing the loss gradient is developed in \secref{secAdj}. \secref{secProps} presents desired properties of the generalization, and our specific choice is given in \secref{secChoice}. Experiment results are presented in \secref{secExperiments}, and we conclude in \secref{secConclusions}. Some of the analysis, and additional experimental results, are deferred to the supplementary material due to lack of space.

\subsection{Related Work} \label{secRelated}
The challenge of choosing the best optimization objective for neural network
training is not a new one. In the past, the quadratic loss was typically used
with gradient-based learning in neural networks \cite{rumelhart1988learning},
but a line of studies demonstrated both theoretically and empirically that the
cross-entropy loss has preferable properties over the quadratic-loss, such as
better learning speed \cite{levin1988accelerated}, better performance
\cite{golik2013cross} and a more suitable shape of the error surface
\cite{glorot2010understanding}. Other cost functions have also been considered. For instance, a novel cost function was proposed in \cite{silva2006new}, but it is not clearly advantageous to cross-entropy. The authors of \cite{bahdanau2015task} address this question in a different setting of sequence prediction. 

Our method allows controlling the sensitivity of the training process to examples with a large prediction bias. When this sensitivity is low, the method can be seen as a form of implicit outlier detection or noise reduction. Several previous works attempt to explicitly remove outliers or noise in neural network training. In one work \cite{smith2011improving}, data is preprocessed to detect label noise induced from overlapping classes, and in another work \cite{jeatrakul2010data} the authors use an auxiliary neural network to detect noisy examples. In contrast, our approach requires a minimal modification on gradient-based training algorithms for neural networks and allows emphasizing examples with a large prediction bias, instead of treating these as noise.

The interplay between ``easy'' and ``hard'' examples during neural network training has been
addressed in the framework of Curriculum Learning
\cite{bengio2009curriculum}. In this framework it is suggested that
training could be more successful if the network is first presented with easy
examples, and harder examples are gradually added to the training process. 
In another work \cite{kumar2010self}, the authors define easy and hard
examples based on the fit to the current model parameters. They propose a
curriculum learning algorithm in which a tunable parameter controls the
proportions of easy and hard examples presented to a learner at each
phase. Our method is simpler than curriculum learning approaches, in that the
examples can be presented at random order to the network. In addition, our
method allows also a heightened sensitivity to harder examples. In a more recent work \cite{zaremba2014learning}, the authors indeed find that a curriculum in which harder examples are presented in early phases outperforms a curriculum that at first uses only easy examples.

%%% Local Variables:
%%% mode: latex
%%% TeX-master: "cost_change"
%%% End:

\newcommand{\err}{\mathrm{err}}
\section{Setting and Notation} \label{secSetting}
For any integer $n$, denote $[n] = \{1,\ldots,n\}$. For a vector $v$, its $i$'th coordinate is denoted $v(i)$. 

We consider a standard feed-forward multilayer neural network \cite{svozil1997introduction}, where the output layer is a softmax layer \cite{bridle1990probabilistic}, with $n$ units, each representing a class. 
Let $\Theta$ denote the neural network parameters, and let $z_j(x;\Theta)$ denote the value of output unit $j$ when the network has parameters $\Theta$, before the applying the softmax function. Applying the softmax function, the probability assigned by the network to class $j$ is $p_j(x;\Theta) := e^{z_j} / \sum\limits_{i=1}^n e^{z_i}$. The label predicted by the network for example $x$ is $\hat{y}(x;\Theta) = \argmax_{j \in [n]} p_j(x;\Theta)$. We consider the task of supervised learning of $\Theta$, using a labeled training sample $S = \{(x_i,y_i)\}_{i=1}^m$, , where $y_i \in [n]$, by optimizing the loss function: $L(\Theta) := \sum\limits_{i=1}^m \ell((x_i,y_i);\Theta)$. A popular choice for $\ell$ is the cross-entropy cost function, defined by $\ell((x,y);\Theta) := -\log p_{y}(x;\Theta)$. 

\section{Generalizing the gradient} \label{secAdj} 
Our proposed method allows controlling the sensitivity of the training procedure to examples on which the network has large errors in prediction, by means of generalizing the gradient. A na{\"i}ve alternative towards the same goal would be using an exponential version of the cross-entropy loss: $\ell = -|\log (p_y)^k|$, where $p_y$ is the probability assigned to the correct class and $k$ is a hyperparameter controlling the sensitivity level. However, the derivative of this function with respect to $p_y$ is an undesired term since it is not monotone in $k$ for a fixed $p_y$, resulting in lack of relevant meaning for small or large values of $k$. The gradient resulting from the above form is of a desired form only for $k=1$, due to cancellation of terms from the derivatives of $l$ and the softmax function. Another na{\"i}ve option would be to consider $l = -\log (p_y^k)$, but this is only a scaled version of the cross-entropy loss and amounts to a change in the learning rate. 

In general, controlling the loss function alone is not sufficient for controlling the relative importance to the training procedure of examples on which the network has large and small errors in prediction. Indeed, when computing the gradients, the derivative of the loss function is being multiplied by the derivative of the softmax function, and the latter is a term that also contains the probabilities assigned by the model to the different classes. Alternatively, controlling the parameters updates themselves, as we describe below, is a more direct way of achieving the desired effect.

Let $(x,y)$ be a single
labeled example in the training set, and consider the partial derivative 
of $\ell(\Theta; (x,y))$ with respect to some parameter $\theta$ in $\Theta$.
We have
\[
\frac{\p \ell((x,y); \Theta)}{\p \theta} = \sum\limits_{j=1}^n 
\frac{\p \ell}{\p z_j}\frac{\p z_j}{\p \theta},
\]
where $z_j$ is the input to the softmax layer when the input example is $x$, and the network parameters are $\Theta$. 
 
If $\ell$ is the cross-entropy loss, we have $\frac{\p \ell}{\p z_j} = \frac{\p \ell}{\p p_y} \frac{\p p_y}{\p z_j}$ and
%$\frac{\p \ell}{\p p_y} = -\frac{1}{p_y}$ 
%and 
%$\frac{\p p_y}{\p z_j} = \begin{cases} p_y (1-p_y) & j=y,\\
%-p_y p_j & j \neq y.
%\end{cases}$
\begin{align*}
\frac{\p \ell}{\p p_y} &= -\frac{1}{p_y},\\
\frac{\p p_y}{\p z_j} &= \begin{cases} p_y (1-p_y) & j=y,\\
-p_y p_j & j \neq y.
\end{cases}
\end{align*}
Hence 
\[
\frac{\p \ell}{\p z_j} = \begin{cases} p_j - 1 & y = j \\
p_j & \text{otherwise}.
\end{cases}
\]
For given $x,y,\Theta$, define the \emph{prediction bias} of the network for example $x$ on class $j$, denoted by $\epsilon_j$, as the (signed) difference between the probability assigned by the network to class $j$ and the probability that should have been assigned, based on the true label of this example. We get $\epsilon_j= p_j-1$ for $j = y$, and $\epsilon_j = p_j$ otherwise. Thus, for the cross-entropy loss,
\begin{equation} \label{eqCE}
\frac{\p \ell}{\p \theta} = \sum\limits_{j=1}^n \frac{\p z_j}{\p \theta} \epsilon_j.
\end{equation}
In other words, when using the cross entropy loss, the effect of any single training example on the gradient is linear in the prediction bias of the current network on this example.

As discussed in \secref{secIntroduction}, it is likely that in many cases, the results of training could be improved if the effect of a single example on the gradient is not linear in the prediction bias. Therefore, we propose a generalization of the gradient that allows non-linear dependence in $\epsilon$. 

For given $x,y,\Theta$ and for $j \in \{1,\ldots,n\}$, define $f:[-1,1]^n \rightarrow \mathbb{R}^n$, let $\epsilon = (\epsilon_1,\ldots,\epsilon_n)$, and consider the following generalization of $\frac{\p \ell}{\p \theta}$:
\begin{equation} \label{eqF}
g(\theta) := \sum\limits_{j=1}^n \frac{\p z_j}{\p \theta} f_j(\epsilon).
\end{equation}
Here $f_j$ is the $j$'th component of $f$. 
When $f$ is the identity, we have $f_j(\epsilon) \equiv \frac{\p \ell}{\p z_j}$, and $g(\theta) =\frac{\p \ell}{\p \theta}$. However, we are now at liberty to study other assignments for $f$. 

We call the vector of values of $g(\theta)$ for $\theta$ in $\Theta$ a \emph{pseudo-gradient}, and propose to use $g$ in place of the gradient within any gradient-based algorithm. In this way, optimization of the cross-entropy loss is replaced by a different algorithm of a similar form. However, as we show in \secref{secExistence}, $g$ is not necessarily the gradient of any loss function.

\section{Properties of $f$}\label{secProps}
Consider what types of functions are reasonable to use for $f$ instead of the identity. First, we expect $f$ to be monotonic non-decreasing, so that a larger prediction bias never results in a smaller update. This is a reasonable requirement if we cannot identify outliers, that is, training examples that have a wrong label. We further expect $f$ to be positive when $j \neq y$ and negative otherwise.  

\begin{figure}[ht]
\begin{center}
\centerline{\includegraphics[width=0.9\columnwidth]{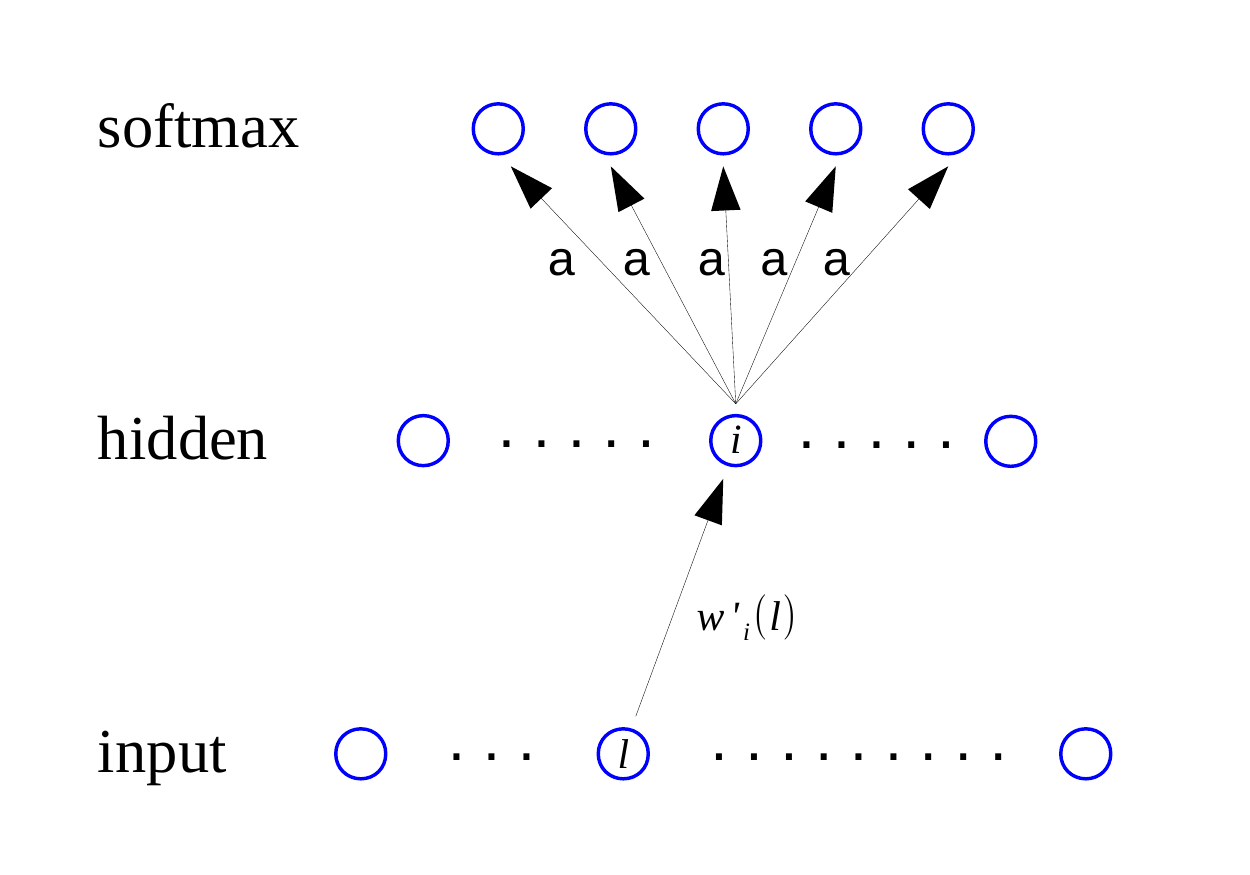}}
\caption{Illustrating the state of the network discussed in above.}
\label{figConstraint}
\end{center}
\end{figure}

In addition to these natural properties, we introduce an additional property that we wish to enforce. To motivate this property, we consider the following simple example. Assume a network with one hidden layer and a softmax layer (see \figref{figConstraint}), where the inputs to the softmax layer are $z_j = \dotprod{w_j,h}+b_j$ and the outputs of the hidden layer are $h(i) = \dotprod{w'_i,x}+b'_i$, where $x$ is the input vector, and $b'_i,w'_i$ are the scalar bias and weight vector between the input layer and the hidden layer. Suppose that at some point during training, hidden unit $i$ is connected to all units $j$ in the softmax layer with the same positive weight $a$. In other words, for all $j \in [n]$, $w_j(i) = a$.  Now, suppose that the training process encounters a training example $(x,y)$, and let $l$ be some input coordinate.

What is the change to the weight $w'_i(l)$ that this training example should cause? Clearly it need not change if $x(l) = 0$, so we consider the case $x(l) \neq 0$. Only the value $h(i)$ is directly affected by changing $w'_i(l)$. 
From the definition of $p_j(x;\Theta)$, the predicted probabilities are fully determined by the ratios $e^{z_j}/e^{z_{j'}}$, or equivalently, by the differences $z_j-z_{j'}$, for all $j,j' \in [n]$. 
Now, $z_j-z_{j'} = \dotprod{w_j,h}+b_j - \dotprod{w_{j'},h}+b_{j'}$. Therefore, $\frac{\p (z_j - z_{j'})}{\p h(i)} = w_j(i) - w_{j'}(i) = a-a = 0$, and therefore 
\[
\frac{\p (z_j - z_{j'})}{\p w'_i(l)} = \frac{\p (z_j - z_{j'})}{\p h(i)} \frac{\p h(i)}{\p w'_i(l)} = 0.
\]
We conclude that in the case of equal weights from unit $i$ to all output units, there is no reason to change the weight $w'_i(l)$ for any $l$. Moreover, preliminary experiments show that in these cases it is desirable to keep the weight stationary, as otherwise it can cause numerical instability due to explosion or decay of weights.

Therefore, we would like to guarantee this behavior also for our pseudo-gradients. Therefore, we require $g(w'_i(l)) = 0$ in this case. It follows that
\begin{align*}
0 &= g(w'_i(l)) = \sum\limits_{j=1}^n \frac{\p z_j}{\p w'_i(l)} f(\epsilon_j) \\
&= \sum\limits_{j=1}^n \frac{\p z_j}{\p h(i)} \frac{\p h(i)}{\p w'_i(l)} f(\epsilon_j) = \sum\limits_{j=1}^n a \cdot x(l) \cdot f(\epsilon_j).\
\end{align*}

Dividing by $a \cdot x(l)$, we get the following desired property for the function $f$, for any vector $\epsilon$ of prediction biases:
\begin{equation}\label{eqConstraint}
f_y(\epsilon) = -\sum \limits_{j \neq y} f_j(\epsilon).
\end{equation}
Note that this indeed holds for the cross-entropy loss, since $\sum_{j\in[n]}\epsilon_j = 0$, and in the case of cross-entropy, $f$ is the identity.

\section{Our choice of $f$}\label{secChoice}
In the case of the cross-entropy, $f$ is the identity, leading to a linear dependence on $\epsilon$. A natural generalization is to consider higher order polynomials. Combining this approach with the requirement in \eqref{eqConstraint}, we get the following assignment for $f$, where $k >0$ is a parameter.

\begin{equation} \label{eqChoice}
f_j(\epsilon) = \begin{cases} -|\epsilon_y|^k &j = y, \\
\frac{|\epsilon_y|^k}{\sum \limits_{i \neq y} \epsilon_i^k} \cdot \epsilon_j^k & \text{otherwise}. \end{cases}
\end{equation}
The expression $\frac{|\epsilon_y|^k}{\sum \limits_{i \neq y} \epsilon_i^k}$ is a normalization term which makes sure \eqref{eqConstraint} is satisfied.
Setting $k = 1$, we get that $g(\theta)$ is the gradient of the cross-entropy loss. Other values of $k$ result in different pseudo-gradients. 

\begin{figure}[ht]
\begin{center}
\centerline{\includegraphics[width=0.8\columnwidth]{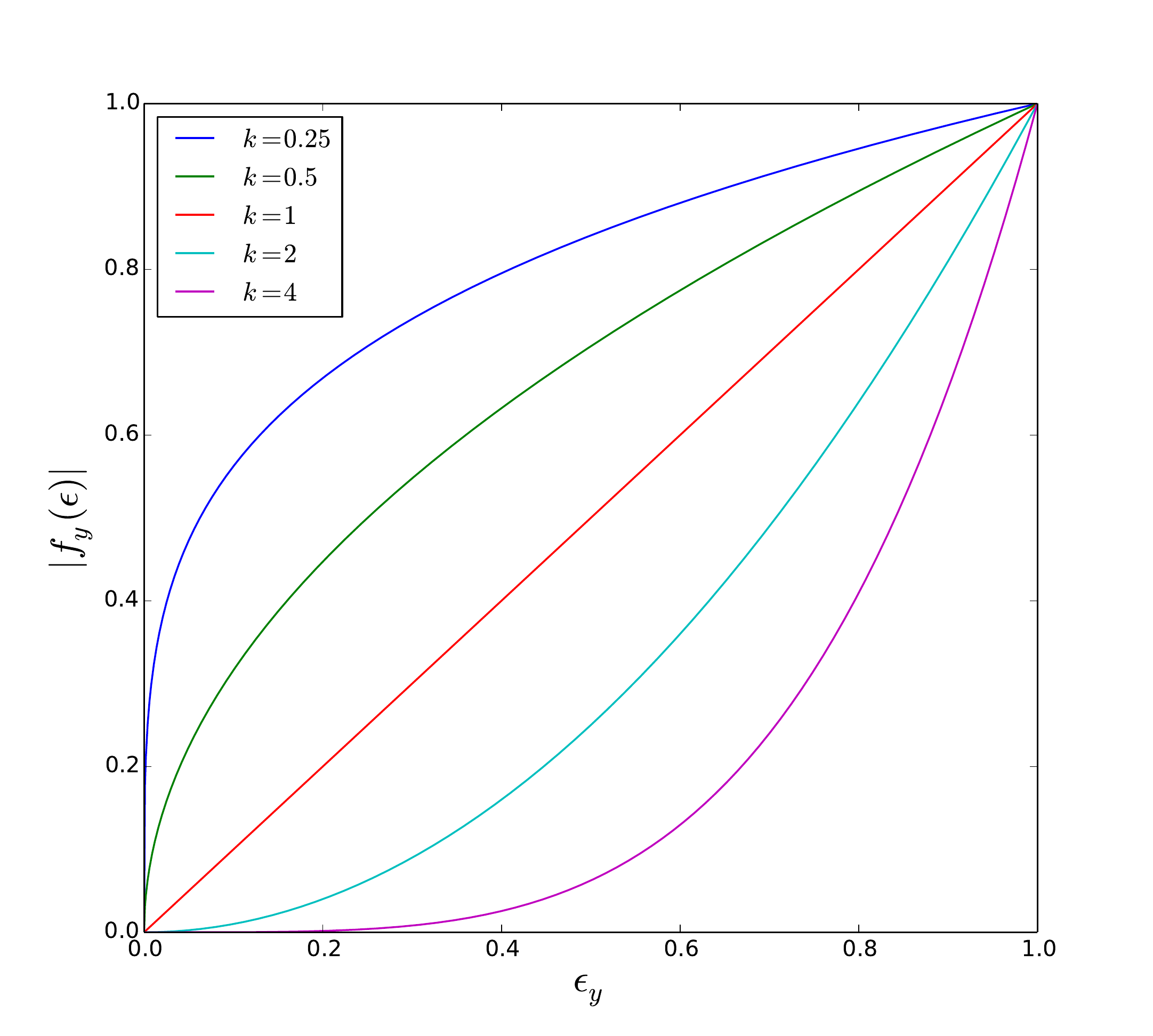}}
\caption{Size of $f_y(\epsilon)$ for different choices of $k$. Lines are in the same order as in the legend.}
%  Size of $f_y(\epsilon}$ for different choices of $k$
\label{figChoices}
\end{center}
\end{figure}

To illustrate the relationship between the value of $k$ and the effect of prediction biases of different sizes on the pseudo-gradient, we plot $f_y(\epsilon)$ as a function of $\epsilon_y$ for several values of $k$ (see \figref{figChoices}). Note that absolute values of the pseudo-gradient are of little importance, since in gradient-based algorithms, the gradient (or in our case, the pseudo-gradient) is usually multiplied by a scalar learning rate which can be tuned. 

As the figure shows, when $k$ is large, the pseudo-gradient is more strongly affected by large prediction biases, compared to small ones. This follows since $\frac{|\epsilon|^k}{|\epsilon'|^k}$ is monotonic increasing in $k$ for $\epsilon > \epsilon'$. On the other hand, when using a small positive $k$ we get that $\frac{|\epsilon|^k}{|\epsilon'|^k}$ tends to $1$, therefore, the pseudo-gradient in this case would be much less sensitive to examples with large prediction biases. Thus, the choice of $f$, parameterized by $k$, allows tuning the sensitivity of the training process to large errors. We note that there could be other reasonable choices for $f$ which have similar desirable properties. We leave the investigation of such other choices to future work.

%Next we discuss, in \secref{secToy}, a very simple toy example which further motivates our choice of $f$, and report some experiments on this example. In \secref{secExistence} we show that the pseudo-gradient defined by $f$ is not the gradient of any cost function. 
\tabone
\subsection{A Toy Example} \label{secToy}
To further motivate our choice of $f$, we describe a very simple example of a distribution and a neural network.
Consider a neural network with no hidden layers, and only one input unit connected to two softmax units. Denoting the input by $x$, the input to softmax unit $i$ is $z_i = w_ix + b_i$, where $w_i$ and $b_i$ are the network weights and biases respectively.

It is not hard to see that the set of possible prediction functions $x \mapsto \hat{y}(x; \Theta)$ that can be represented by this network is exactly the set of threshold functions of the form $\hat{y}(x;\Theta) = \sign(x-t)$ or $\hat{y}(x; \Theta) = -\sign(x-t)$. 

For convenience assume the labels mapped to the two softmax units are named $\{-1,+1\}$. 
Let $\alpha \in (\frac{1}{2},1)$, and suppose that labeled examples are drawn independently at random from the following distribution $\cD$ over $\reals \times \{-1,+1\}$: Examples are uniform in $[-1,1]$; Labels of examples in $[0,\alpha]$ are deterministically $1$, and they are $-1$ for all other examples. For this distribution, the prediction function with the smallest prediction error that can be represented by the network is $x\mapsto \sign(x)$. 

However, optimizing the cross-entropy loss on the distribution, or in the limit of a large training sample, would result in a different threshold, leading to a larger prediction error (for a detailed analysis see \appref{apToy} in the supplementary material). Intuitively, this can be traced to the fact that the examples in $(\alpha,1]$ cannot be classified correctly by this network when the threshold is close to $0$, but they still affect the optimal threshold for the cross-entropy loss. 

Thus, for this simple case, there is motivation to move away from optimizing the cross-entropy, to a different update rule that is less sensitive to large errors. This reduced sensitivity is achieved by our update rule with $k < 1$. On the other hand, larger values of $k$ would result in higher sensitivity to large errors, thereby degrading the classification accuracy even more. 

We thus expect that when training the network using our new update rule, the prediction error of the resulting network should be monotonically increasing with $k$, hence values of $k$ which are smaller than $1$ would give a smaller error. 
We tested this hypothesis by training this simple network on a synthetic dataset generated according to the distribution $\cD$ described above, with $\alpha = 0.95$.

We generated 30,000 examples for each of the training, validation and test datasets. The biases were initialized to 0 and the weights were initialized from a uniform distribution on $(-0.1,0.1)$. We used batch gradient descent with a learning rate of $0.01$ for optimization of the four parameters, where the gradient is replaced with the pseudo-gradient from \eqref{eqF}, using the function $f$ defined in \eqref{eqChoice}. $f$ is parameterized by $k$, and we performed this experiment using values of $k$ between $0.0625$ and $4$. After each epoch, we computed the prediction error on the validation set, and training was stopped after 3000 epochs in which this error was not changed by more than $0.001\%$. The values of the parameters at the end of training were used to compute the misclassification rate on the test set.

\tabref{tabToy} reports the results for these experiments, averaged over $10$ runs for each value of $k$. The results confirm our hypothesis regarding the behavior of the network for the different values of $k$, and further motivate the possible benefits of using $k \neq 1$. Note that while the prediction error is monotonic in $k$ in this experiment, the cross-entropy is not, again demonstrating the fact that optimizing the cross-entropy is not optimal in this case. 

\begin{table}[t]
\caption{Toy example experiment results}
\vskip 0.15in
\begin{center}
\begin{small}
%\begin{sc}
\begin{tabular}{cccc}
%\abovespace\belowspace
\toprule
$k$ & Test error & Threshold & CE Loss\\
\midrule
%\abovespace\
4		& 8.36\% & 0.116 & 0.489\\
2		& 6.73\% & 0.085 & 0.361\\
1		& 4.90\% & 0.049 & 0.288\\
0.5		& 4.27\% & 0.037 & 0.299\\
0.25		& 4.04\% & 0.030 & 0.405\\
0.125	& 3.94\% & 0.028 & 0.625\\
0.0625	& 3.61\% & 0.022 & 1.190\\
\bottomrule
%\belowspace

\end{tabular}
%\end{sc}
\end{small}
\end{center}
%\vskip -0.1in
\label{tabToy}
\end{table}

%%% Local Variables:
%%% mode: latex
%%% TeX-master: "cost_change_sivan"
%%% End:

\subsection{Non-existence of a Cost Function for $f$} \label{secExistence}
It is natural to ask whether, with our choice of $f$ in \eqref{eqChoice}, $g(\theta)$ is the gradient of another cost function, instead of the cross-entropy. The following lemma demonstrates that this is not the case. 

\begin{lemma}
Assume $f$ as in \eqref{eqChoice} with $k \neq 1$, and $g(\Theta)$ the resulting pseudo-gradient. There exists a neural network for which the $g(\Theta)$ is not a gradient of any cost function.
\end{lemma}

The proof of is lemma is left for the supplemental material. Note that the above lemma does not exclude the possibility that a gradient-based algorithm that uses $g$ instead of the gradient still somehow optimizes some cost function.

%%% Local Variables:
%%% mode: latex
%%% TeX-master: "cost_change_sivan"
%%% End:

%
%
\section{Experiments}\label{secExperiments}
%\tabonemnist
%\tabonesvhn
%\tabonecifarten
%\tabonecifarhundred
%\tabone
\tabthree
\tabfive
%\tabonecifarten
%\tabonecifarhundred

For our experiments, we used four classification benchmark datasets from the field of computer vision: The MNIST dataset \cite{lecun1998gradient}, the Street View House Numbers dataset (SVHN) \cite{netzer2011reading} and the CIFAR-10 and CIFAR-100 datasets \cite{krizhevsky2009learning}. A more detailed description of the datasets can be found in \appref{apExp} in the supplementary material.

The neural networks we experimented with are feed-forward neural networks that contain one, three or five hidden layers of various layer sizes. For optimization, we used stochastic gradient descent with momentum \cite{sutskever2013importance} with several values of momentum and a minibatch size of 128 examples. For each value of $k$, we replaced the gradient in the algorithm with the pseudo-gradient from \eqref{eqF}, using the function $f$ defined in \eqref{eqChoice}. For the multilayer experiments we also used Gradient-Clipping \cite{DBLP:conf/icml/PascanuMB13} with a threshold of 100. In the hidden layers, biases were initialized to 0 and for the weights we used the initialization scheme from \cite{glorot2010understanding}. Both biases and weights in the softmax layer were initialized to 0. %\footnote{A link to the test code will be provided in the non-anonymous version of this paper.}

In each experiment, we used cross-validation to select the best value of $k$. The learning rate was optimized using cross-validation for each value of $k$ separately, as the size of the pseudo-gradient can be significantly different between different values of $k$, as evident from \eqref{eqChoice}. We compared the test error between the models using the selected $k$ and $k=1$, each with its best performing learning rate. Additional details about the experiment process can be found in \appref{apSelect} in the supplementary material.

We report the test error of each of the trained models for MNIST, SVHN, CIFAR-10 and CIFAR-100 in Tables  \ref{tabOneLayer},  \ref{tab3layer} and \ref{tab5layer} for networks with one, three and five layers respectively. Additional experiments are reported in \appref{apExp} in the supplementary material. We further report the cross-entropy values using the selected $k$ and the default $k=1$. 

Several observations are evident from the experiment results. First, aligned with our hypothesis, the value of $k$ selected by the cross-validation scheme was almost always smaller than $1$, for the shallow networks, larger than one for the deep networks, and close to one for  networks with medium depth. Indeed, the capacity of network is positively correlated with the optimal sensitivity to hard examples. 

Second, for the shallow networks the cross-entropy loss on the test set was always worse for the selected $k$ than for $k=1$. This implies that indeed, by using a different value of $k$ we are not optimizing the cross-entropy loss, yet are improving the success of optimizing the true prediction error. On the contrary, in the experiments with three and five layers, the cross entropy is also improved by selecting the larger $k$. This is an interesting phenomenon, which might be explained by the fact that examples with a large prediction bias have a high cross-entropy loss, and so focusing training on these examples reduces the empirical cross-entropy loss, and therefore also the true cross-entropy loss. 

%
%First, for one-layer network, in almost all the experiments (\tabref{tabOneLayer}), the value of $k$ selected by the cross-validation scheme was smaller than $1$, which resulted in a lower test error in almost all experiments. This is aligned with our hypothesis, that in a network with a limited capacity, it might be better to train with less sensitivity to hard examples. Moreover, the cross-entropy loss on the test set was always worse for the selected $k$ than for $k=1$. This implies that indeed, by using a different value of $k$ we are not optimizing the cross-entropy loss, yet are improving the success of optimizing the true prediction error. 
%
%Second, for networks with 3 layers (\tabref{tab3layer}), the most common value selected by the cross-validation scheme was $1$, while for 5 layers (\tabref{tab5layer}), in most experiments the selected value of $k$ was larger than $1$. This is again aligned with the assumption that as the capacity of the network grows, higher sensitivity to hard examples is useful. In the experiments with 3 and 5 layers, unlike the case of a single layer, the cross entropy is also improved by selecting the larger $k$. This is an interesting phenomenon, which might be explained by the fact that examples with a large prediction bias have a high cross-entropy loss, and so focusing training on these examples reduces the empirical cross-entropy loss, and therefore also the true cross-entropy loss. 

To summarize, our experiments show that overall, cross-validating over the value of $k$ usually yields improved results over $k=1$, and that, as expected, the optimal value of $k$ grows with the depth of the network.

%%% Local Variables:
%%% mode: latex
%%% TeX-master: "cost_change_sivan"
%%% End:

\section{Conclusions}\label{secConclusions}
Inspired by an intuition in human cognition, in this work we proposed a generalization of the cross-entropy gradient step in which a tunable parameter controls the sensitivity of the training process to hard examples. Our experiments show that, as we expected, the optimal level of sensitivity to hard examples is positively correlated with the depth of the network. Moreover, the experiments demonstrate that selecting the value of the sensitivity parameter using cross validation leads overall to improved prediction error performance on a variety of benchmark datasets. 

The proposed approach is not limited to feed-forward neural networks --- it can be used in any gradient-based training algorithm, and for any network architecture. In future work, we plan to study this method as a tool for improving training in other architectures, such as convolutional networks and recurrent neural networks, as well as experimenting with different levels of sensitivity to hard examples in different stages of the training procedure, and combining the predictions of models with different levels of this sensitivity. 

\section*{Acknowledgments}
This work has been supported by the European Community’s Seventh Framework Programme through the ERC Starting Grant No. 338164 (iHEARu). Sivan Sabato was supported in part by the Israel Science Foundation (grant No. 555/15).

%%% Local Variables:
%%% mode: latex
%%% TeX-master: "cost_change_sivan"
%%% End:

\bibliographystyle{aaai} 
\bibliography{cost_change}

\begin{thebibliography}{}

\bibitem[\protect\citeauthoryear{Bahdanau \bgroup et al\mbox.\egroup
  }{2015}]{bahdanau2015task}
Bahdanau, D.; Serdyuk, D.; Brakel, P.; Ke, N.~R.; Chorowski, J.; Courville, A.;
  and Bengio, Y.
\newblock 2015.
\newblock Task loss estimation for sequence prediction.
\newblock {\em arXiv preprint arXiv:1511.06456}.

\bibitem[\protect\citeauthoryear{Bengio \bgroup et al\mbox.\egroup
  }{2009}]{bengio2009curriculum}
Bengio, Y.; Louradour, J.; Collobert, R.; and Weston, J.
\newblock 2009.
\newblock Curriculum learning.
\newblock In {\em Proc. of the 26th annual International Conference on Machine
  Learning (ICML)},  41--48.
\newblock Montreal, Canada: ACM.

\bibitem[\protect\citeauthoryear{Bridle}{1990}]{bridle1990probabilistic}
Bridle, J.~S.
\newblock 1990.
\newblock Probabilistic interpretation of feedforward classification network
  outputs, with relationships to statistical pattern recognition.
\newblock In {\em Neurocomputing}. Springer.
\newblock  227--236.

\bibitem[\protect\citeauthoryear{Cho, Courville, and
  Bengio}{2015}]{cho2015describing}
Cho, K.; Courville, A.; and Bengio, Y.
\newblock 2015.
\newblock Describing multimedia content using attention-based encoder-decoder
  networks.
\newblock {\em IEEE Transactions on Multimedia} 17(11):1875--1886.

\bibitem[\protect\citeauthoryear{Glorot and
  Bengio}{2010}]{glorot2010understanding}
Glorot, X., and Bengio, Y.
\newblock 2010.
\newblock Understanding the difficulty of training deep feedforward neural
  networks.
\newblock In {\em Proc. of International Conference on Artificial Intelligence
  and Statistics},  249--256.

\bibitem[\protect\citeauthoryear{Golik, Doetsch, and
  Ney}{2013}]{golik2013cross}
Golik, P.; Doetsch, P.; and Ney, H.
\newblock 2013.
\newblock Cross-entropy vs. squared error training: a theoretical and
  experimental comparison.
\newblock In {\em Proc. of INTERSPEECH},  1756--1760.

\bibitem[\protect\citeauthoryear{Hinton \bgroup et al\mbox.\egroup
  }{2012}]{hinton2012deep}
Hinton, G.; Deng, L.; Yu, D.; Dahl, G.~E.; Mohamed, A.-r.; Jaitly, N.; Senior,
  A.; Vanhoucke, V.; Nguyen, P.; Sainath, T.~N.; et~al.
\newblock 2012.
\newblock Deep neural networks for acoustic modeling in speech recognition: The
  shared views of four research groups.
\newblock {\em Signal Processing Magazine, IEEE} 29(6):82--97.

\bibitem[\protect\citeauthoryear{Hinton}{1989}]{hinton1989connectionist}
Hinton, G.~E.
\newblock 1989.
\newblock Connectionist learning procedures.
\newblock {\em Artificial intelligence} 40(1):185--234.

\bibitem[\protect\citeauthoryear{Jeatrakul, Wong, and
  Fung}{2010}]{jeatrakul2010data}
Jeatrakul, P.; Wong, K.~W.; and Fung, C.~C.
\newblock 2010.
\newblock Data cleaning for classification using misclassification analysis.
\newblock {\em Journal of Advanced Computational Intelligence and Intelligent
  Informatics} 14(3):297--302.

\bibitem[\protect\citeauthoryear{Keren and Schuller}{2016}]{Keren16-CRAa}
Keren, G., and Schuller, B.
\newblock 2016.
\newblock Convolutional {RNN}: an enhanced model for extracting features from
  sequential data.
\newblock In {\em {Proc. of 2016 International Joint Conference on Neural
  Networks (IJCNN)}},  3412--3419.

\bibitem[\protect\citeauthoryear{Keren \bgroup et al\mbox.\egroup
  }{2016}]{keren2016convolutional}
Keren, G.; Deng, J.; Pohjalainen, J.; and Schuller, B.
\newblock 2016.
\newblock Convolutional neural networks with data augmentation for classifying
  speakers’ native language.
\newblock In {\em Proc. of INTERSPEECH},  2393--2397.

\bibitem[\protect\citeauthoryear{Kingma and Ba}{2015}]{kingma2015adamICLR}
Kingma, D., and Ba, J.
\newblock 2015.
\newblock Adam: A method for stochastic optimization.
\newblock In {\em International Conference on Learning Representations (ICLR)}.

\bibitem[\protect\citeauthoryear{Krizhevsky and
  Hinton}{2009}]{krizhevsky2009learning}
Krizhevsky, A., and Hinton, G.
\newblock 2009.
\newblock Learning multiple layers of features from tiny images.

\bibitem[\protect\citeauthoryear{Krizhevsky, Sutskever, and
  Hinton}{2012}]{krizhevsky2012imagenet}
Krizhevsky, A.; Sutskever, I.; and Hinton, G.~E.
\newblock 2012.
\newblock Imagenet classification with deep convolutional neural networks.
\newblock In {\em Proc. of Advances in Neural Information Processing Systems
  (NIPS)},  1097--1105.

\bibitem[\protect\citeauthoryear{Kumar, Packer, and
  Koller}{2010}]{kumar2010self}
Kumar, M.~P.; Packer, B.; and Koller, D.
\newblock 2010.
\newblock Self-paced learning for latent variable models.
\newblock In {\em Proc. of Advances in Neural Information Processing Systems
  (NIPS)},  1189--1197.

\bibitem[\protect\citeauthoryear{Lake \bgroup et al\mbox.\egroup
  }{2016}]{lake2016building}
Lake, B.~M.; Ullman, T.~D.; Tenenbaum, J.~B.; and Gershman, S.~J.
\newblock 2016.
\newblock Building machines that learn and think like people.
\newblock {\em arXiv preprint arXiv:1604.00289}.

\bibitem[\protect\citeauthoryear{LeCun \bgroup et al\mbox.\egroup
  }{1998}]{lecun1998gradient}
LeCun, Y.; Bottou, L.; Bengio, Y.; and Haffner, P.
\newblock 1998.
\newblock Gradient-based learning applied to document recognition.
\newblock {\em Proceedings of the IEEE} 86(11):2278--2324.

\bibitem[\protect\citeauthoryear{Levin and
  Fleisher}{1988}]{levin1988accelerated}
Levin, E., and Fleisher, M.
\newblock 1988.
\newblock Accelerated learning in layered neural networks.
\newblock {\em Complex systems} 2:625--640.

\bibitem[\protect\citeauthoryear{Netzer \bgroup et al\mbox.\egroup
  }{2011}]{netzer2011reading}
Netzer, Y.; Wang, T.; Coates, A.; Bissacco, A.; Wu, B.; and Ng, A.~Y.
\newblock 2011.
\newblock Reading digits in natural images with unsupervised feature learning.
\newblock In {\em NIPS workshop on deep learning and unsupervised feature
  learning}.
\newblock Granada, Spain.

\bibitem[\protect\citeauthoryear{Pascanu, Mikolov, and
  Bengio}{2013}]{DBLP:conf/icml/PascanuMB13}
Pascanu, R.; Mikolov, T.; and Bengio, Y.
\newblock 2013.
\newblock On the difficulty of training recurrent neural networks.
\newblock In {\em Proceedings of the 30th International Conference on Machine
  Learning (ICML)},  1310--1318.

\bibitem[\protect\citeauthoryear{Polyak}{1964}]{polyak1964some}
Polyak, B.~T.
\newblock 1964.
\newblock Some methods of speeding up the convergence of iteration methods.
\newblock {\em USSR Computational Mathematics and Mathematical Physics}
  4(5):1--17.

\bibitem[\protect\citeauthoryear{Rumelhart, Hinton, and
  Williams}{1988}]{rumelhart1988learning}
Rumelhart, D.~E.; Hinton, G.~E.; and Williams, R.~J.
\newblock 1988.
\newblock Learning representations by back-propagating errors.
\newblock {\em Cognitive modeling} 5:3.

\bibitem[\protect\citeauthoryear{Silva \bgroup et al\mbox.\egroup
  }{2006}]{silva2006new}
Silva, L.~M.; De~Sa, J.~M.; Alexandre, L.; et~al.
\newblock 2006.
\newblock New developments of the {Z-EDM} algorithm.
\newblock In {\em Intelligent Systems Design and Applications}, volume~1,
  1067--1072.

\bibitem[\protect\citeauthoryear{Smith and Martinez}{2011}]{smith2011improving}
Smith, M.~R., and Martinez, T.
\newblock 2011.
\newblock Improving classification accuracy by identifying and removing
  instances that should be misclassified.
\newblock In {\em The 2011 International Joint Conference on Neural Networks
  (IJCNN)},  2690--2697.

\bibitem[\protect\citeauthoryear{Sutskever \bgroup et al\mbox.\egroup
  }{2013}]{sutskever2013importance}
Sutskever, I.; Martens, J.; Dahl, G.; and Hinton, G.
\newblock 2013.
\newblock On the importance of initialization and momentum in deep learning.
\newblock In {\em Proc. of the 30th International Conference on Machine
  Learning (ICML)},  1139--1147.

\bibitem[\protect\citeauthoryear{Sutskever, Vinyals, and
  Le}{2014}]{sutskever2014sequence}
Sutskever, I.; Vinyals, O.; and Le, Q.~V.
\newblock 2014.
\newblock Sequence to sequence learning with neural networks.
\newblock In {\em Advances in neural information processing systems},
  3104--3112.

\bibitem[\protect\citeauthoryear{Svozil, Kvasnicka, and
  Pospichal}{1997}]{svozil1997introduction}
Svozil, D.; Kvasnicka, V.; and Pospichal, J.
\newblock 1997.
\newblock Introduction to multi-layer feed-forward neural networks.
\newblock {\em Chemometrics and intelligent laboratory systems} 39(1):43--62.

\bibitem[\protect\citeauthoryear{Szegedy \bgroup et al\mbox.\egroup
  }{2015}]{Szegedy_2015_CVPR}
Szegedy, C.; Liu, W.; Jia, Y.; Sermanet, P.; Reed, S.; Anguelov, D.; Erhan, D.;
  Vanhoucke, V.; and Rabinovich, A.
\newblock 2015.
\newblock Going deeper with convolutions.
\newblock In {\em The IEEE Conference on Computer Vision and Pattern
  Recognition (CVPR)}.

\bibitem[\protect\citeauthoryear{Tieleman and
  Hinton}{2012}]{tieleman2012lecture}
Tieleman, T., and Hinton, G.
\newblock 2012.
\newblock Lecture 6.5-rmsprop: Divide the gradient by a running average of its
  recent magnitude.
\newblock {\em COURSERA: Neural Networks for Machine Learning}.

\bibitem[\protect\citeauthoryear{Zaremba and
  Sutskever}{2014}]{zaremba2014learning}
Zaremba, W., and Sutskever, I.
\newblock 2014.
\newblock Learning to execute.
\newblock {\em arXiv preprint arXiv:1410.4615}.

\end{thebibliography}

\clearpage
\begin{center}
\hrule
{\Large{\ourtitle}}

\vspace{0.5em}
{\large{Supplementary material}}
\hrule
\end{center}

\appendix

\newcommand{\ce}{\mathrm{CE}}
\section{Proof for Toy Example}\label{apToy}
Consider the neural network from the toy example in \secref{secToy}. In this network, there exists one classification threshold such that examples above or below it are classified to different classes. We prove that for a large enough training set, the value of the cross-entropy cost is not minimal when the threshold is at $0$. 

Suppose that there is an assignment of network parameters that minimizes the cross-entropy which induces a threshold at $0$. 
The output of the softmax layer is determined uniquely by $\frac{e^{z_0}}{e^{z_1}}$, or equivalently by $z_0 - z_1 = x(w_0-w_1) + b_0 - b_1$. Therefore, we can assume without loss of generality that $w_1 = b_1 = 0$. Denote $w:=w_0, b:=b_0$. If $w=0$ in the minimizing assignment, then all examples are classified as members of the same class and in particular, the classification threshold is not zero. Therefore we may assume $w\neq 0$. In this case, the classification threshold is $\frac{-b}{w}$. Since we assume a minimal solution at zero, the minimizing assignment must have $b=0$. 

%We have a uniform distribution on $[-1,-1]$, therefore with density of $\frac{1}{2}$. Recall that $\alpha \in (0,1)$. 
When the training set size approaches infinity, the cross-entropy on the sample approaches the expected cross-entropy on $\cD$. Let $\ce(w,b)$ be the expected cross-entropy on $\cD$ for network parameter values $w,b$. Then
\begin{align*}
\ce(w,b) = -\frac{1}{2}\big(&\int_{-1}^0 \log(p_0(x))dx + \int_0^\alpha \log (p_1(x))dx \\ 
&+ \int_\alpha^1 \log (p_0(x))dx\big).
\end{align*}
And we have:
\begin{align*}
\log(p_0(x)) &= \log \frac{e^{wx + b}}{e^{wx + b}+1} = wx+b - \log(e^{wx+b}+1),\\
\log(p_1(x)) &= \log \frac{1}{e^{wx + b}+1}  = - \log(e^{wx+b}+1).
\end{align*}
Therefore
\begin{align*}
&\frac{\p\ce(w,b)}{\p b} =\\
&-\frac{1}{2} \frac{\p }{\p b}\Big(
 \int_{-1}^0 (wx+b)dx
- \int_{-1}^0  \log(e^{wx+b}+1)dx \\
&\qquad\qquad - \int_0^\alpha   \log(e^{wx+b}+1)dx \\ 
&\qquad\qquad + \int_\alpha^1 (wx +b)dx 
- \int_\alpha^1  \log(e^{wx+b}+1)dx
\Big)
\end{align*}
\[ = -\frac{1}{2} (
1
- \frac{\p }{\p b} \left(\int_{-1}^1  \log(e^{wx+b}+1)dx \right)
+ 1-\alpha). \]
Differentiating under the integral sign, we get
\begin{align*}
\frac{\p\ce(w,b)}{\p b} = -\frac{1}{2} \left(2-\alpha -\int_{-1}^1 \frac{e^{wx+b}}{e^{wx+b} + 1}\right)
\end{align*}
Since we assume the cross-entropy has a minimal solution with $b=0$, we have
\begin{align*}
0 &= -2\frac{\p\ce(w,b=0)}{\p b}\\ 
  &=2-\alpha -\frac{1}{w}(\log(e^{w}+1) - \log(e^{-w}+1)).
\end{align*}
Therefore
\[ 
w(2-\alpha) = \log \frac{e^{w}+1}{e^{-w}+1} = \log( e^w) = w.
\]
Since $\alpha \neq 1$, it must be that $w = 0$. This contradicts our assumption, hence the cross-entropy does not have a minimal solution with a threshold at $0$.

\section{Proof of Lemma 1}
\begin{proof}
Consider a neural network with three units in the output layer, and at least one hidden layer.
Let $(x,y)$ be a labeled example, and suppose that there exists some cost function $\bar\ell((x,y);\Theta)$, differentiable in $\Theta$, such that for $g$ as defined in \eqref{eqF} and $f$ defined in \eqref{eqChoice} for some $k > 0$, we have $g(\theta) = \frac{\p \bar\ell}{\p \theta}$ for each parameter $\theta$ in $\Theta$. We now show that this is only possible if $k=1$. 

Under the assumption on $\bar \ell$, for any two parameters $\theta_1,\theta_2$, 
\[
\frac{\p}{\p \theta_2}\left(\frac{\p \bar\ell}{\p \theta_1}\right) = \frac{\p^2\bar\ell}{\p \theta_1 \theta_2} = \frac{\p}{\p \theta_1}\left(\frac{\p \bar\ell}{\p \theta_2}\right),
\]
hence
\begin{equation}\label{eqEqual}
\frac{\p g(\theta_1)}{\p \theta_2}  = \frac{\p g(\theta_2)}{\p \theta_1}.
\end{equation}

Recall our notations: $h(i)$ is the output of unit $i$ in the last hidden layer before the softmax layer, $w_j(i)$ is the weight between the hidden unit $i$ in the last hidden layer, and unit $j$ in the softmax layer, $z_j$ is the input to unit $j$ in the softmax layer, and $b_j$ is the bias of unit $j$ in the softmax layer. 
%Recall the derivatives of the softmax activation function mentioned in \secref{secSetting}: $\frac{\p p_j}{\p z_j} = p_j(1-p_j)$ and for $i \neq j$: $\frac{\p p_j}{\p z_i} = -p_j p_i$. 

Let $(x,y)$ such that $y = 1$. From \eqref{eqEqual} and \eqref{eqF}, we have
\[
\frac{\p}{\p w_2(1)} \sum \limits_{j=1}^n \frac{\p z_j}{\p w_1(1)} f_j(\epsilon) = \frac{\p}{\p w_1(1)} \sum \limits_{j=1}^n \frac{\p z_j}{\p w_2(1)} f_j(\epsilon).
\]

Plugging in $f$ as defined in \eqref{eqChoice}, and using the fact that $\frac{\p z_j}{\p w_i(1)} = 0$ for $i \neq j$, we get:
\begin{align*}
&-\frac{\p}{\p w_2(1)} \left(\frac{\p z_1}{\p w_1(1)}\cdot|\epsilon_1|^k\right) =\\
&\qquad \frac{\p}{\p w_1(1)} \left(\frac{\p z_2}{\p w_2(1)} \cdot \frac{\epsilon_2^k}{\epsilon_2^k + \epsilon_3^k}\cdot |\epsilon_1|^k\right).
\end{align*}
Since $y=1$, we have $\epsilon = (p_1-1, p_2,p_3)$. In addition, 
$\frac{\p z_j}{\p w_j(1)} = h(1)$ and $\frac{\p h(1)}{\p w_j(1)} = 0$ for $j \in [2]$. Therefore
\begin{align}\label{eqGrad}
&-\frac{\p}{\p w_2(1)} \left((1-p_1)^k\right) = \\
&\qquad\frac{\p}{\p w_1(1)} \left( \frac{p_2^k}{p_2^k + p_3^k} (1-p_1)^k\right).\notag
\end{align}

Next, we evaluate each side of the equation separately, using the following: 
\begin{align*}
&\frac{\p p_j}{\p w_j(1)} = \frac{\p p_j}{\p z_j}\frac{\p z_j}{\p w_{j}(1)}  = h(1)p_j(1-p_j),\\
\forall j \neq i,\quad &\frac{\p p_j}{\p w_i(1)} =  \frac{\p p_j}{\p z_i}\frac{\p z_i}{\p w_i(1)} = -h(1)p_ip_j.
\end{align*}

For the LHS of \eqref{eqGrad}, we have
\begin{align*}
-\frac{\p}{\p w_2(1)} (1-p_1)^k &= -k(1-p_1)^{k-1}h(1)p_1p_2.
\end{align*}
For the RHS, 
\iffalse
starting with the numerator:
\begin{align*}
    \frac{\p}{\p w_1(1)} p_2^k (1-p_1)^k =
%    &-kp_2^{k-1}h(1)p_1p_2 (1-p_1)^k & \\ 
%    & -p_2^k k(1-p_1)^{k-1} h(1)p_1(1-p_1) =& \\
    -2kh(1)p_1p_2^k(1-p_1)^k.
\end{align*}

And the denominator: 
\begin{align*}
    &\frac{\p}{\p w_1(1)} (p_2^k + p_3^k) =
%    &-kp_2^{k-1}h(1)p_2p_1 -kp_3^{k-1}h(1)p_3p_1 = 
-kh(1)p_1(p_2^k + p_3^k).
\end{align*}
\fi 
\begin{align*}
    \frac{\p}{\p w_1(1)}  \frac{p_2^k}{p_2^k + p_3^k} (1-p_1)^k =  
%    &\frac{1}{(p_2^k+p_3^k)^2} (-2kh(1)p_1p_2^k(1-p_1)^k(p_2^k + p_3^k) \\ 
%    & + h(1)p_2^k(1-p_1)^k kp_1(p_2^k+p_3^k))=& \\
    -\frac{kh(1)p_1p_2^k(1-p_1)^k}{p_2^k+p_3^k}.
\end{align*}

Hence \eqref{eqGrad} holds if and only if:
\[ 
1 = \frac{p_2^{k-1}(1-p_1)}{p_2^k+p_3^k}.
\]
For $k=1$, this equality holds since $p_1+p_2+p_3 = 1$. However, for any $k \neq 1$, there are values of $p_1,p_2,p_3$ such that this does not hold. We conclude that our choice of $f$ does not lead to a pseudo-gradient $g$ which is the gradient of any cost function.
\end{proof}

\section{Additional Experiment details and Results}
\subsection{datasets} \label{apExp}
The MNIST dataset \cite{lecun1998gradient}, consisting of grayscale 28x28 pixel images of handwritten digits, with 10 classes, 60,000 training examples and 10,000 test examples, the Street View House Numbers dataset (SVHN) \cite{netzer2011reading}, consisting of RGB 32x32 pixel images of digits cropped from house numbers, with 10 classes 73,257 training examples and 26,032 test examples and the CIFAR-10 and CIFAR-100 datasets \cite{krizhevsky2009learning}, consisting of RGB 32x32 pixel images of 10/100 object classes, with 50,000 training examples and 10,000 test examples. All datasets were linearly transformed such that all features are in the interval $[-1,1]$.

\subsection{Choosing the value of $k$} \label{apSelect}
In each experiment, we used cross-validation to select the best value of $k$. For networks with one hidden layer, $k$ was selected out of the values $\{4,2,1,0.5,0.25,0.125,0.0625\}$. For networks with 3 or 5 hidden layers, $k$ was selected out of the values $\{4,2,1,0.5,0.25\}$, removing the smaller values of $k$ due to performance considerations (in preliminary experiments, these small values yielded poor results for deep networks). The learning rate was optimized using cross-validation for each value of $k$ separately, as the size of the pseudo-gradient can be significantly different between different values of $k$, as evident from \eqref{eqChoice}.

For each experiment configuration, defined by a dataset, network architecture and momentum, we selected an initial learning rate $\eta$, based on preliminary experiments on the training set. Then the following procedure was carried out for $\eta/2,\eta,2\eta$, for every tested value of $k$: 
\begin{enumerate}
\item Randomly split the training set into 5 equal parts, $S_1,\ldots,S_5$.
\item Run the iterative training procedure on $S_1\cup S_2 \cup S_3$, until there is no improvement in test prediction error for $15$ epochs on the early stopping set, $S_4$.
\item Select the network model that did the best on $S_4$.
\item Calculate the validation error of the selected model on the validation set, $S_5$.
\item Repeat the process for $t$ times after permuting the roles of $S_1,\ldots,S_5$. We set $t=10$ for MNIST, and $t=7$ for CIFAR-10/100 and SVHN.
\item Let $\err_{k,\eta}$ be the average of the $t$ validation errors.
\end{enumerate}
We then found $\argmin_{\eta} \err_{k,\eta}$. If the minimum was found with the minimal or the maximal $\eta$ that we tried, we also performed the above process using half the $\eta$ or double the $\eta$, respectively. This continued iteratively until there was no need to add learning rates. At the end of this process we selected $(k^*,\eta^*) = \argmin_{k,\eta} \err_{k,\eta}$, and retrained the network with parameters $k^*,\eta^*$ on the training sample, using one fifth of the sample as an early stopping set. We compared the test error of the resulting model to the test error of a model retrained in the same way, except that we set $k =1$ (leading to standard cross-entropy training), and the learning rate to $\eta_1^* = \argmin_{\eta}\err_{1,\eta}$. The final learning rates in the selected models were in the range $[10^{-1},10]$ for MNIST, and $[10^{-4},1]$ for the other datasets. 

\subsection{Results} \label{apTables}
Additional experiment results with momentum values other than $0.5$ are reported in \tabref{tabOneLayerSup}, \tabref{tab3layerSup}.
\taboneSup
\tabthreeSup

\end{document}